\documentclass[runningheads]{llncs}
\usepackage{orcidlink}
\usepackage[T1]{fontenc}
\usepackage{booktabs}
\usepackage{tabularx}
\usepackage{makecell}
\usepackage{multirow}
\usepackage{array}
\usepackage{caption}
\usepackage{subcaption}
\usepackage{graphicx}

\usepackage{booktabs}
\usepackage{color}

\begin{document}
\title{Effect of User-Prompted Priors on Semi-Automated Cancer Lesion Segmentation in Whole-Body Computed Tomography}
\titlerunning{User-Prompted Priors for Semi-Automated Cancer Lesion Segmentation}
% If the paper title is too long for the running head, you can set
% an abbreviated paper title here
%
\author{Isac Stark\inst{1,2,3}\orcidlink{0009-0005-8484-8467}\email{isac.stark@uu.se} \and Johan Öfverstedt\inst{1}\orcidlink{0000-0003-0253-9037} \and Elin Lundström\inst{1}\orcidlink{0000-0003-2955-4958} \and Simon Ekström\inst{3}\orcidlink{0000-0003-2368-6888} \and Håkan Ahlström\inst{1,3}\orcidlink{0000-0002-8701-969X} \and Joel Kullberg\inst{1,2,3}\orcidlink{0000-0001-8205-7569}}

\authorrunning{I. Stark et al.}
% First names are abbreviated in the running head.
% If there are more than two authors, 'et al.' is used.
%
\institute{Department of Surgical Sciences, Uppsala University, Sweden \and Department of Surgical Sciences, SciLifeLab, Uppsala University, Sweden \and Antaros Medical, Mölndal, Sweden\\}
\maketitle              % typeset the header of the contribution
\begin{abstract}
    In clinical oncology studies, metastatic cancer is commonly evaluated using \emph{Response Evaluation Criteria in Solid Tumors} (RECIST), in which the diameter of up to five lesions is measured and followed over the course of treatment. However, RECIST shows limited correlation with overall survival. Total tumour volume (TTV) is a stronger predictor but typically relies on manual ground-truth segmentation of all lesions, which is time-consuming and requires expert domain knowledge. Semi-automated approaches leveraging user-prompted priors, such as bounding boxes and single-slice contours, as inputs to automated segmentation methods can facilitate the generation of ground-truth segmentations. This work investigates the impact of different user-prompted priors on semi-automated cancer lesion segmentation performance in whole-body computed tomography. Across 3-fold cross-validation and external testing, more complex spatial priors consistently improved performance, with contour priors from three orthogonal planes (axial, coronal and sagittal) achieving the best results. On the external test (n=3865 lesions), this approach achieved a mean Dice score of $0.882$, compared to a mean Dice score of $0.671$ for the baseline model with no spatial prior. These findings suggest that the use of multi-plane orthogonal user-prompted priors can improve semi-automated tumour lesion segmentation and support efficient generation of high-quality volumetric ground-truth data.

\end{abstract}
\section{Introduction}
Cancer is one of the leading causes of death worldwide, \cite{brayGlobalCancerStatistics2024} and the majority of cancer-related deaths are in turn attributed to metastatic cancer \cite{hudockFutureTrendsIncidence2023}. In clinical metastatic oncology trials, \emph{overall survival} (OS) is regarded as the gold standard endpoint for evaluating treatment response \cite{delgadoClinicalEndpointsOncology2021}. However, studies relying on this endpoint generally require large cohorts and long study durations to achieve statistical significance, increasing both costs and time-to-market for new cancer treatments \cite{saadStatisticalControversiesClinical2016}.

Instead, imaging-based surrogate endpoints such as \emph{progression-free survival} (PFS) or \emph{objective response rate} (ORR) are commonly used \cite{mittalFrequentlyAskedQuestions2024}. While OS is a patient-centric endpoint focusing on the outcome of the individual, PFS and ORR measure how the disease itself is progressing, which enables earlier evaluation of treatment response. These surrogate endpoints are usually derived via the \emph{Response Evaluation Criteria in Solid Tumours} (RECIST 1.1) \cite{eisenhauerNewResponseEvaluation2009}, in which the diameter of up to five target lesions are measured during the course of treatment. The sum of these diameters and their percentage change is then considered as a proxy of the overall tumour burden and treatment response. However, while these surrogate endpoints have been used extensively \cite{kornOverviewProgressionFreeSurvival2013,aykanObjectiveResponseRate2020} in clinical oncology trials, they have shown limited correlation with OS, and even been associated with a decreased OS for certain subpopulations \cite{merinoIrreconcilableDifferencesDivorce2023}. To measure the PFS or ORR, computed tomography (CT) \cite{pirastehImagingItsImpact2021} is the most commonly used imaging modality due to its low cost and wide availability; this modality is also recommended according to the RECIST guidelines.

A promising and more comprehensive alternative to uni-dimensional measurements, such as RECIST, is \emph{total tumour volume} (TTV), in which the volume of all cancerous lesions is quantified \cite{zeeuwTotalTumorVolume2025}. Manually delineating each tumour in metastatic cancer is, however, a time-consuming task subject to high inter-observer variability \cite{huangInterobserverVariabilityTumor2018}, making it infeasible in large-scale clinical trials. An alternative to manual delineation is the use of either fully or semi-automatic segmentation methods. However, malignant lesions exhibit a high degree of inter- and intra-tumour heterogeneity \cite{ganeshanQuantifyingTumourHeterogeneity2013}, which increases the demand for large volumes of high-quality ground-truth data required to train fully automatic deep learning-based segmentation methods. Due to this, semi-automatic  segmentation methods have gained increasing traction as a compromise between fully automatic approaches and labour-intensive manual delineation, particularly in medical imaging where substantial domain expertise remains of high importance for clinical validity. Semi-automatic methods use input from an operator to improve the end result. This can for example be a spatial prior, such as a bounding box, which defines a coarse region in which the object being segmented exists. Efficient and high-performing semi-automatic segmentation has the added benefit of being intrinsically quality-controlled when performed by a skilled operator, thereby enabling more efficient clinical trials. This would also enable downstream analyses that depend on accurate segmentations, such as radiomics-based feature extraction \cite{mariottiInsightsRadiomicsComprehensive2025} or lesion tracking. Evaluating what constitutes an effective prior for lesion segmentation also opens opportunities for future cascade pipelines with lesion detection and lesion segmentation treated  as two separate steps.

This work investigates the effect of incorporating multi-plane orthogonal slices as priors to improve volumetric segmentation. By varying the number and orientation of orthogonal planes (axial, coronal, sagittal) used to guide the segmentation process, we evaluate how these additional spatial cues influence the accuracy of the resulting 3D segmentations in whole-body CT, with the aim of generating high-quality ground-truth data for future research.

\section{Background \& Related Work}
In parallel, multiple methods have been proposed to improve 3D medical segmentation by leveraging 2D priors across multiple planes. In 2025 Du et al. \cite{duSegVolUniversalInteractive2025} released SegVol, which utilises spatial and semantic encoders to improve general medical image segmentation for CT images. Furthermore, they investigated how their method and several other contemporary methods \cite{kirillovSegmentAnything2023,sunMedicalImageAnalysis2024,wangSAMMed3DGeneralPurposeSegmentation2025,maSegmentAnythingMedical2024} performed on tumour lesion segmentation by evaluating the methods on the DeepLesion3D \cite{yanDeepLesionAutomatedMining2018} dataset from the ULS23 \cite{degrauwULS23ChallengeBaseline2025} challenge. The highest performance was achieved by MedSAM with a median Dice score of $0.768$, followed by their own method at $0.7065$. Recently, Machado et al. introduced ONCOPILOT \cite{machadoPromptableCTFoundation2025}, which extended the SAM approach to tumour-specific segmentation by utilizing 2D priors (points, point edits, and bounding boxes) in combination with autoregressive propagation to generate volumetric segmentations. ONCOPILOT achieves an mean Dice score of $0.70$ for their point and bounding box mode and a mean Dice of $0.78$ for their point-edit mode.

Around the same time, Isensee et al. \cite{isenseeNnInteractiveRedefining3D2025} adapted the highly popular nnU-Net \cite{isenseeNnUNetSelfconfiguringMethod2021} framework for interactive segmentation by incorporating user prompts such as points, bounding boxes, and lasso-based refinements. Their approach focuses on iterative interaction in 2D or slice-wise settings, but does not explicitly study the effect of 3D priors or orthogonal multi-plane guidance for volumetric segmentation.
\section{Materials \& Methods}
\subsection{Dataset}
In total, three publicly available datasets were used, two for training and the third one for external testing to validate the performance and generalizability of the models. For training, the previously mentioned \emph{DeepLesion3D} \cite{yanDeepLesionAutomatedMining2018} dataset, as well as a sub-sample of the \emph{FDG-PET-CT-Lesions} dataset \cite{gatidisResultsAutoPETChallenge2024} was used. The \emph{Longitudinal CT} \cite{kustnerLongitudinalCT2025} cohort was used as test-set. Information on the three datasets is summarized in Table \ref{tab:datasets}. Ethical approval was obtained from the Swedish Ethical Review Authority to conduct research on these retrospective datasets.

\begin{table}[h]
    \centering
    \caption{Overview of the datasets used in this study.}
    \label{tab:datasets}
    \setlength{\tabcolsep}{5pt}
    \renewcommand{\arraystretch}{1.2}
    \resizebox{\textwidth}{!}{%
        \begin{tabular}{l|c|c|c}
            \toprule
                         & DeepLesion3D                                      & FDG-PET-CT-Lesions                   & Longitudinal CT \\
            \midrule
            Modality     & CT                                                & PET/CT                               & CT              \\
            No. Lesions  & 743                                               & 163                                  & 3865            \\
            Cancer types & Mixed                                             & Lymphoma, Melanoma                   & Melanoma        \\
            \multirow{2}{*}{Avg. voxel size}
                         & \multirow{2}{*}{$(0.84 \times 0.84 \times 3.01)$}
                         & PET: $(2.04 \times 2.04 \times 3)$
                         & \multirow{2}{*}{$(0.86 \times 0.86 \times 2.98)$}                                                          \\
                         &                                                   & CT: $(0.87 \times 0.87 \times 1.96)$ &                 \\
            \bottomrule
        \end{tabular}}
\end{table}
\subsubsection{DeepLesion3D:}
The \emph{DeepLesion3D} data contains 743 fully segmented cancer lesions of various types, pre-cropped to a volume of interest and extracted from the larger \emph{DeepLesion} dataset \cite{yanDeepLesionAutomatedMining2018} as part of the ULS23 challenge  \cite{degrauwULS23ChallengeBaseline2025}.
\subsubsection{FDG-PET-CT-Lesions:}
The subset of the FDG-PET-CT-Lesions data used in this work consisted of 39 patients with either lymphoma or melanoma (19 and 20 patients, respectively). The FDG-PET-CT-Lesions dataset was originally released as part of the autoPET challenge \cite{gatidisResultsAutoPETChallenge2024}. In this dataset, ground-truth tumour segmentation was performed according to the metabolic uptake from the fluorodeoxyglucose positron emission tomography (FDG-PET) scan.

Due to potential anatomical misalignment between PET and CT, the ground-truth tumour segmentations were manually adjusted to improve alignment with the anatomy in the CT images. Only segmentations with a visible lesion in CT were corrected and used, excluding areas of possible inflamed tissue or other regions of high FDG uptake. The segmentations were adjusted by four radiologists with 24, 14, 8 and 4 years of experience and reviewed by two other radiologists with 45 and 3.5 years of experience.

Due to the labour-intensive process of correcting the PET segmentations to align with the CT, the adjustments were performed in the slice spacing of the PET-images and thereafter linearly interpolated to intermediate slices in the CT images.
\subsubsection{Longitudinal CT:}
To evaluate the performance and generalizability of the models, especially taking into account that part of the training data is a lower resolution (i.e., the tumour segmentations in FDG-PET-CT-Lesions), the \emph{Longitudinal CT} dataset \cite{kustnerLongitudinalCT2025} was chosen as a holdout test-set. The cohort contains longitudinal scans from 300 patients with melanoma. All reference segmentations were manually delineated by two experienced radiologists \cite{kustnerLongitudinalCT2025}.
\subsection{Method}
\begin{figure}
    \begin{center}
        \includegraphics[width=0.9\linewidth]{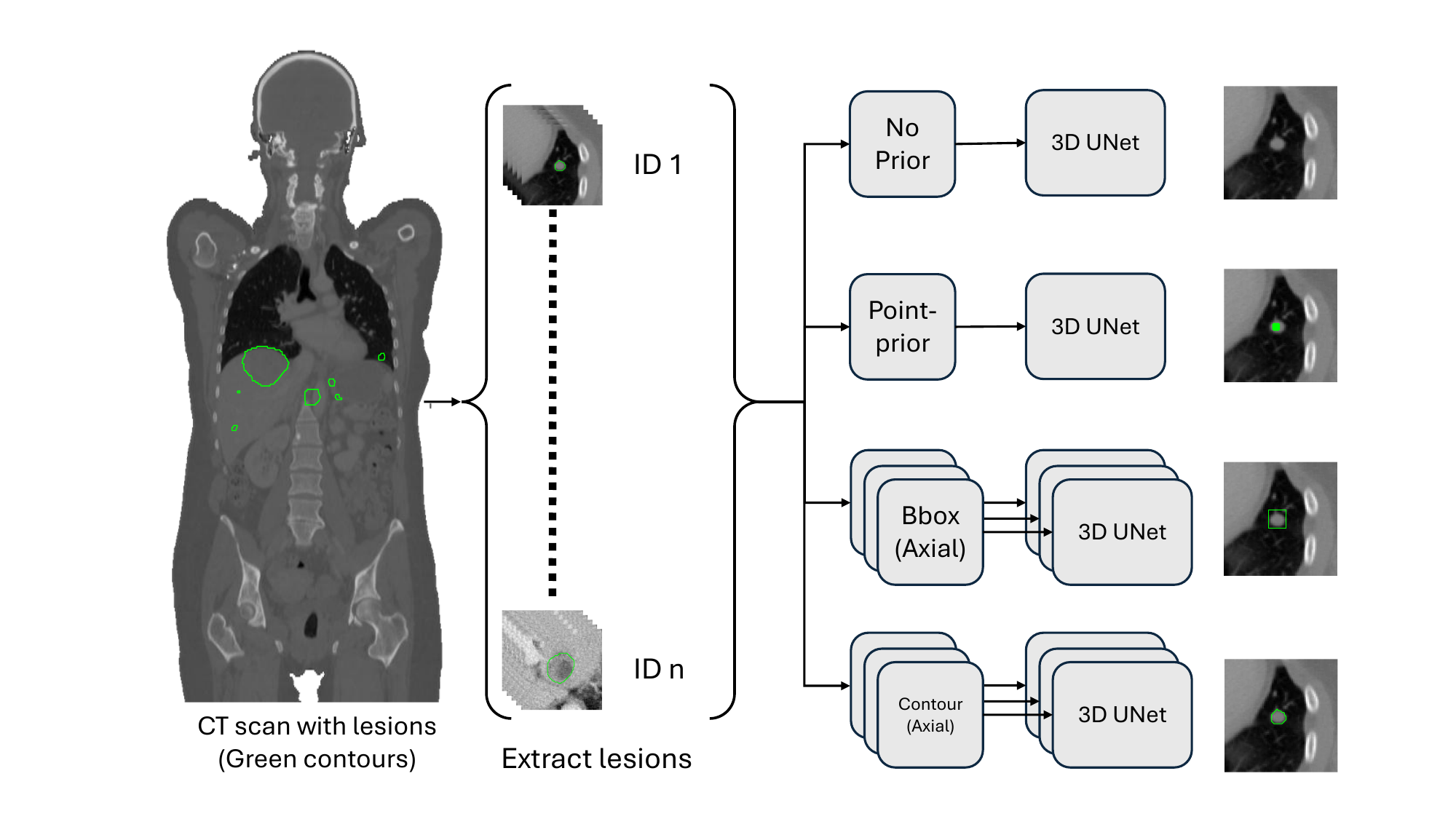}
        \caption{Schematic overview of the proposed method. Lesions are first extracted and cropped from a larger imaging volume. For each lesion, spatial priors are generated and used to train a 3D U-Net under different prior configurations. For the bounding box and contour priors, three variants are trained based on the number of orthogonal slices used (axial, axial+coronal, axial+coronal+sagittal). The point prior has been dilated for visualization purposes.}
        \label{fig:SchemaForPaper}
    \end{center}
\end{figure}

The proposed method consists of three primary steps: (1) lesion cropping, (2) generation of user-prompted priors, and (3) training of a 3D-UNet-based segmentation model using the simulated user-prompted priors. Figure \ref{fig:SchemaForPaper} shows a schematic overview of the entire process.
\subsubsection{Lesion Cropping:}
As an initial pre-processing step, the lesions from both the training and test-set were cropped around their respective center using a fixed volume size. As this work explores how priors affect the performance in user-prompted segmentation, the location of the lesion itself is already known and surrounding anatomy relatively far from the lesion is of limited relevance. The size of the crop was chosen as the $99^{th}$ percentile of size of the lesions in the training set to ensure that the vast majority of the lesions were fully included while still maintaining a standardized volume size. This resulted in a cubic volume with a side length of approximately $6.7\mbox{ cm}$. Lesions exceeding this size were handled using sliding-window inference.
\subsubsection{User-Prompted Priors:}
Three different types of user-prompted priors of varying spatial information (and user effort) were simulated: points, bounding boxes, and contours. For comparison, a version without any prior, but still cropped around the lesion, was also evaluated.

Both the bounding box and contour priors were generated from the ground-truth lesion masks by identifying the slice with the largest lesion cross-section. This was performed either for a single plane (axial), two orthogonal planes (axial and coronal) or all three planes (axial, coronal and sagittal). Increasing the number of anatomical planes increases the spatial information provided to the network, but also the complexity of input for the user to generate. For the bounding boxes, the smallest possible fitting rectangle over the original lesion mask was extracted. For the contour, the slice with the maximal area of the lesion mask itself was extracted along the specified axis.
\subsection{Network architecture \& Segmentation Priors}
We employ the \emph{nnU-Net} \cite{isenseeNnUNetSelfconfiguringMethod2021} method as the backbone for all segmentation experiments. \emph{nnU-Net} is a self-configuring semantic segmentation method that adapts certain parameters in the preprocessing, network architecture and postprocessing based on the fingerprint of the dataset. Recently, nnU-Net has become highly popular as a baseline model in segmentation challenges. By relying on this widely used standardized framework, we ensure that the change in performance can be attributed to the priors used rather than architectural decisions.

All priors were encoded as 3D binary mask volumes and provided as an additional input channel to the nnU-Net pipeline. No normalization scheme was applied to the prior-mask. Two custom training strategies were implemented to simulate the user-generated priors. For the point-prior, a custom trainer was used to both generate the prior itself on the fly for the training, but also for the evaluation. The point-priors were generated by randomly sampling voxels within the lesion mask during training. The number of points were drawn from a Poisson distribution with a fixed offset of one according to the equation below
\begin{equation}
    N \sim \mbox{Pois(0.5)}+1.
\end{equation}
Following this process, the majority of training cases had 1-2 points, while still allowing for additional points to be seen during the training. This sampling was applied as the first step in the data augmentation pipeline. To evaluate the performance of the model during the course of the training, a centroid point-prior was used as part of the validation data augmentation pipeline to maintain consistency for internal validation through reduced randomness.

As bounding boxes and contours were generated to closely match the ground-truth tumour masks, a separate trainer was employed to simulate user-induced variation by introducing offsets in bounding box and contour positions relative to the tumours. As a first step in the training augmentation, the priors were randomly translated from their original position within the range of $\pm3$ voxels for in-plane translations (x- and y-directions) and $\pm1$ voxel for inter-slice translations (z-direction). The rationale for choosing a smaller translation range along the z-direction is the comparably low spatial resolution in this direction (Table \ref{tab:datasets}). Applying larger shifts along the z direction increases the risk of the prior not overlapping with the true lesion, thereby unnecessarily penalizing the network.

For the training, 3-fold cross-validation was performed on the combined training cohort ($n = 906$). This was stratified on lesion size and dataset of origin to give an even spread of cases in all the folds of the training.
\subsection{Statistical Evaluation}
Segmentation performance was evaluated with the Dice score. To evaluate whether increased number of orthogonal slices for the bounding box and contour priors led to a statistically significant improvement, a Wilcoxon signed-rank test was applied to both the cross-validation folds and the test set. The significance level was adjusted using the Bonferroni correction to account for the number of paired samples in the cross-validation ($n=906$) and the test-set ($n=3865$). A  p-value $< \frac{0.05}{n}$ was considered statistically significant. The same statistical test was also applied to compare the two best-performing priors for the bounding boxes and contours.
\section{Results}
\begin{figure}
    \centering
    \begin{subfigure}{.5\textwidth}
        \centering
        \includegraphics[width=1\linewidth]{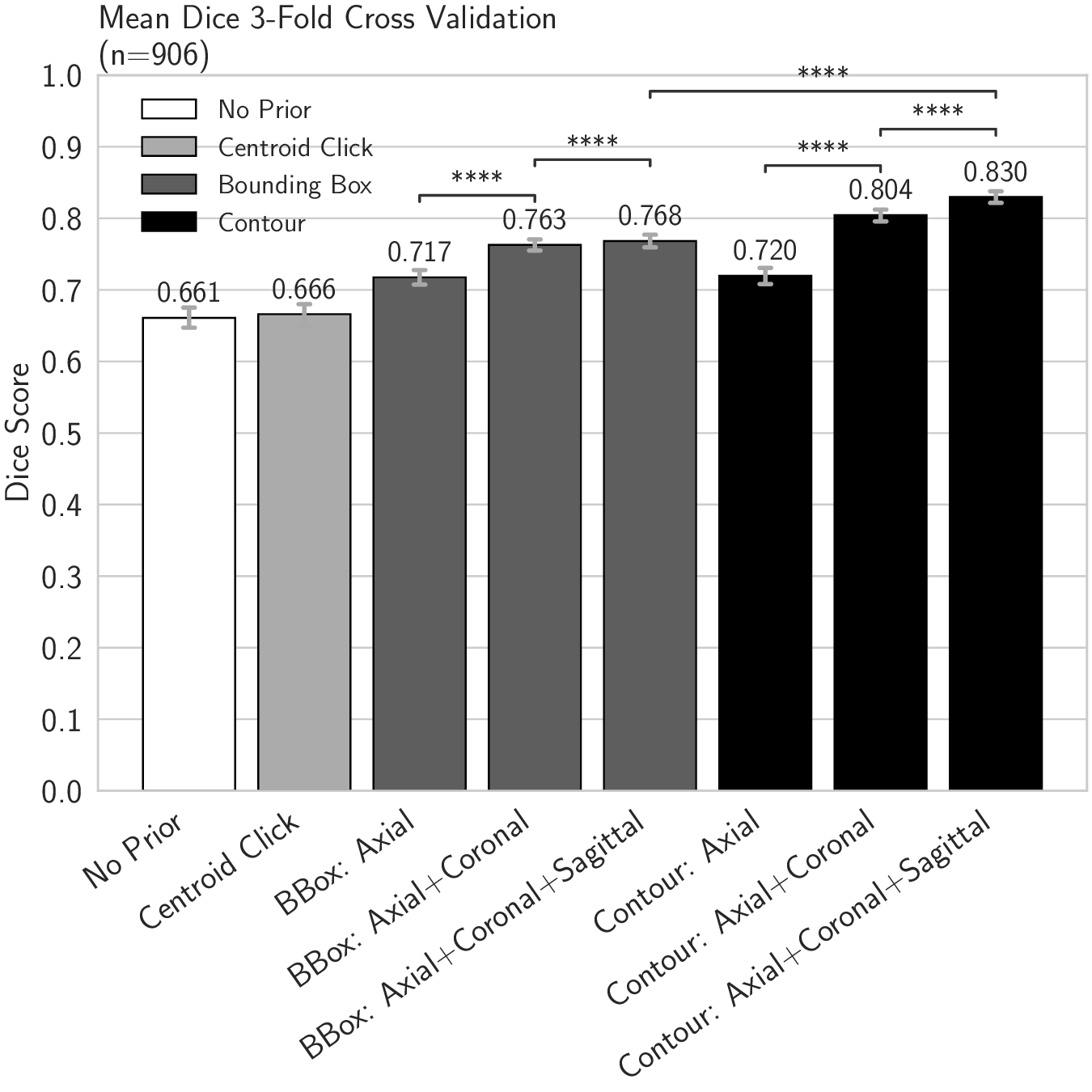}
        \caption{}
        \label{fig:3CV}
    \end{subfigure}%
    \begin{subfigure}{.5\textwidth}
        \centering
        \includegraphics[width=1\linewidth]{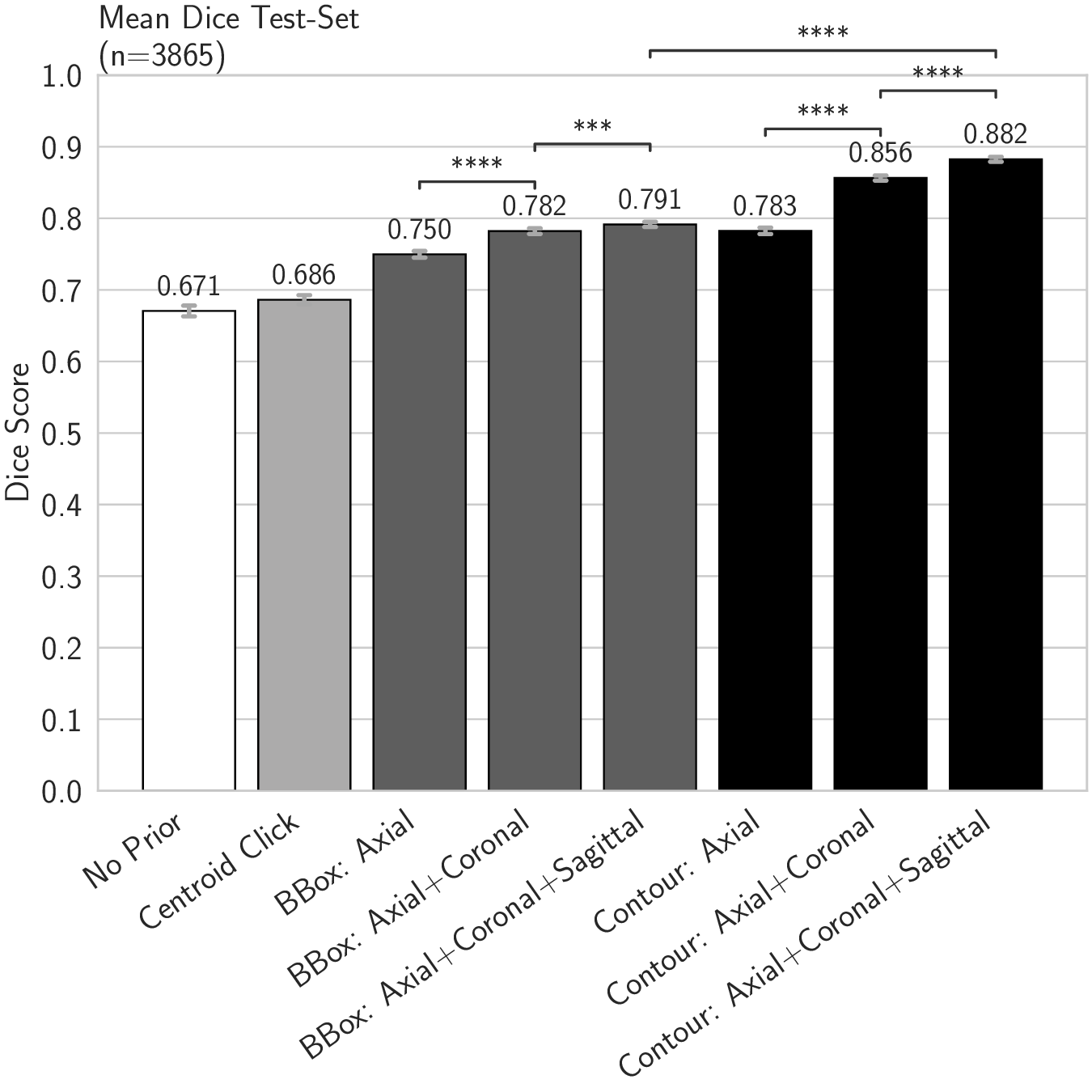}
        \caption{}
        \label{fig:testset}
    \end{subfigure}
    \caption{Mean Dice score across (a) 3-fold cross-validation ($n = 906$) and (b) the external test set ($n = 3865$) for nnU-Net models trained with different spatial priors (no prior, centroid click, bounding box, contour) and varying numbers of orthogonal input slices (axial, coronal, and sagittal). Error bars indicate 95\% confidence intervals. Statistical significance is shown for comparisons between slice configurations within each prior type and between the best-performing models across prior categories (***: $1.00\times10^{-4} < p \leq 1.00\times10^{-3}$; ****: $p \leq 1.00\times10^{-4}$).}
    \label{fig:test}
\end{figure}

Figures \ref{fig:3CV} and \ref{fig:testset} present the mean Dice score across 3-fold cross-validation ($n=906$) and the test-set ($n=3865$) for the nnU-Net models trained with different spatial priors (No prior, centroid click, bounding box, contour) and varying numbers of orthogonal input slices. For both the cross-validation and the test set, the contour prior with multiple orthogonal slices showed the highest performance, achieving a mean Dice of $0.830$ in cross-validation and $0.882$ on the external test set (Fig. \ref{fig:test}). The performed Wilcoxon signed-rank test shows that the performance gained when adding further orthogonal priors for the bounding box and contour priors is statistically significant. Further, the difference in performance between the two best performing priors was significant, even after correcting for multiple comparison.

In Table \ref{tab:perturbation_results} the performance of the bounding box and contour configurations and their variations after perturbation is presented. After perturbation, the performance of all priors decreased, while still remaining relatively high.
\begin{table}[!h]
    \caption{Performance of the various slice configurations in terms of average Dice-score on the test-set ($n=3865$) when allowing the originally generated prior to be translated at random within the specified bounds. In-slice translation is denoted with x and y and inter-slice translation with z.}
    \centering
    \small
    \setlength{\tabcolsep}{4pt}
    \renewcommand{\arraystretch}{1.2}
    \resizebox{\textwidth}{!}{\begin{tabularx}{\textwidth}{l|*{6}{>{\centering\arraybackslash}X}}
            \toprule
            \textbf{Perturbation}
                                        & \rotatebox{-0}{\shortstack{BBox                                                              \\Ax}}
                                        & \rotatebox{-0}{\shortstack{BBox                                                              \\Ax+Cor}}
                                        & \rotatebox{-0}{\shortstack{BBox                                                              \\Ax+Cor\\+Sag}}
                                        & \rotatebox{-0}{\shortstack{Contour                                                           \\Ax}}
                                        & \rotatebox{-0}{\shortstack{Contour                                                           \\Ax+Cor}}
                                        & \rotatebox{-0}{\shortstack{Contour                                                           \\Ax+Cor\\+Sag}} \\
            \midrule
            No Perturbation             & 0.750                               & 0.782 & 0.791 & 0.783 & 0.856          & \textbf{0.882} \\
            $\pm 0 z, \pm 1 x, \pm 1 y$ & 0.743                               & 0.776 & 0.785 & 0.755 & 0.815          & \textbf{0.834} \\
            $\pm 1 z, \pm 1 x, \pm 1 y$ & 0.735                               & 0.765 & 0.770 & 0.744 & \textbf{0.796} & 0.787          \\
            $\pm 1 z, \pm 3 x, \pm 3 y$ & 0.729                               & 0.756 & 0.763 & 0.738 & \textbf{0.792} & 0.790          \\
            \bottomrule
        \end{tabularx}}
    \label{tab:perturbation_results}
\end{table}

\section{Discussion \& Conclusions}
We have performed a large scale evaluation on how different user-prompted priors affect the performance of semi-automatic tumour segmentation in whole-body CT. The results show that generally, the more detailed the spatial prior, the higher the performance, with contour-based prompts in three orthogonal planes achieving the highest Dice score before and after translation of the prior. The progressively improved performance observed when moving from no prior to simple point prompts, and further to bounding boxes and contour priors, indicates that more detailed initial priors help guide the network toward more accurate tumour segmentations. The contour-based priors were especially effective, most likely as they not only provide information about the extent of the lesion, but also a boundary to non-tumour surrounding tissue. Whereas the bounding box only provides a rough region of interest.

Another clear trend across both the cross-validation and the test set was that the inclusion of additional orthogonal planes in the prior resulted in higher mean Dice scores. This is expected, as cancerous lesions may exhibit irregular shapes that are not fully captured by a single anatomical plane, and due to more information being provided to the model by the user. The difference in performance between the different priors and the number of orthogonal priors highlights an important trade-off between segmentation accuracy and user effort. The simpler priors, such as the points or a single plane bounding box, may be faster for an operator to generate, and may provide acceptable performance for lesions with a clearly defined boundary or a homogenous appearance, while a more complex prior might, be necessary for lesions with low contrast to surrounding healthy tissue.

Another important aspect of this work is generalizability. Both the training and test data contain lesions from patients suffering from melanoma, however, there is still a wide variety of lesions in both, which suggests that there is some degree of generalizability to the method. It remains to be seen how this extends to other forms of cancer. In fact, all priors performed slightly better on the test set. This is most likely due to the fact that the ground-truth segmentations from the FDG-PET-CT-Lesions dataset used in the training were originally created at a relatively coarse PET voxel size $(2.04 \times 2.04 \times 3)$ compared to the CT image resolution employed for rest of the training data.

Limitations of this study include the use of priors simulated from ground-truth segmentations rather than generated by operators. Therefore, annotation time, real-world clinical usability, and the impact of operator variability on performance cannot be directly assessed. The experiments were performed with a single segmentation backbone (nnU-Net), meaning that conclusions may depend on the chosen architecture, which was out of scope by design. Lastly, the work does not address lesion detection, where most of the time is spent while segmenting the total tumour volume.

Overall, the study indicates that user-prompted tumour segmentation can benefit from richer spatial priors, especially when these priors provide information from multiple orthogonal planes. The results suggest that contour-based guidance is particularly effective, but that the optimal prior may, in practice, depend on the balance between annotation effort and segmentation accuracy. Future work should therefore investigate real interactive use, quantify annotation time savings, and explore whether different lesion types benefit from different prompting strategies. Such studies would help clarify the most appropriate integration of user-prompted segmentation into practical radiological workflows.

\begin{credits}
    \subsubsection{\ackname}
    This work was supported by the SciLifeLab \& Wallenberg Data Driven Life Science Program (grant: KAW2024.0159), the Swedish Cancer Society under Grant 23 3123 Pj, Lions Cancerforskningsfond and G och E Erikssons stiftelse för cancerforskning.

    \subsubsection{\discintname}
    IS, SE, HA and JK are employees of Antaros Medical.
\end{credits}
%
% ---- Bibliography ----
%
% BibTeX users should specify bibliography style 'splncs04'.
% References will then be sorted and formatted in the correct style.
%
\bibliographystyle{splncs04}
\bibliography{Paper1_manualPriors}
\end{document}